\documentclass[10pt,twocolumn,letterpaper]{article}

 \usepackage[pagenumbers]{iccv} 
\usepackage[accsupp]{axessibility}  
\definecolor{iccvblue}{rgb}{0.21,0.49,0.74}
\usepackage[pagebackref,breaklinks,colorlinks,allcolors=iccvblue]{hyperref}
\usepackage{algorithm}
\usepackage{algpseudocode}
\usepackage{amsmath}
\usepackage{tcolorbox} 
\usepackage{graphicx}
\usepackage{booktabs}
\usepackage{adjustbox}
\usepackage{xcolor}
\usepackage{float}
\usepackage{wrapfig}
\usepackage{amsthm}

\def\paperID{8426} 
\def\confName{ICCV}
\def\confYear{2025}

\title{Grouped Speculative Decoding for Autoregressive Image Generation }

\author{
Junhyuk So$^{1}$ \quad
Juncheol Shin$^{2}$ \quad
Hyunho Kook$^{1}$ \quad
Eunhyeok Park$^{1,2}$ \\
$^{1}$Department of Computer Science and Engineering, POSTECH\\
$^{2}$Graduate School of Artificial Intelligence, POSTECH\\
{\tt\small \{junhyukso,jchshin, kookhh0827, eh.park\}@postech.ac.kr}
}

\begin{document}

\twocolumn[{%
\renewcommand\twocolumn[1][]{#1}%
\maketitle
\begin{center}
    \centering
    \captionsetup{type=figure}
    \includegraphics[width=0.99\textwidth]{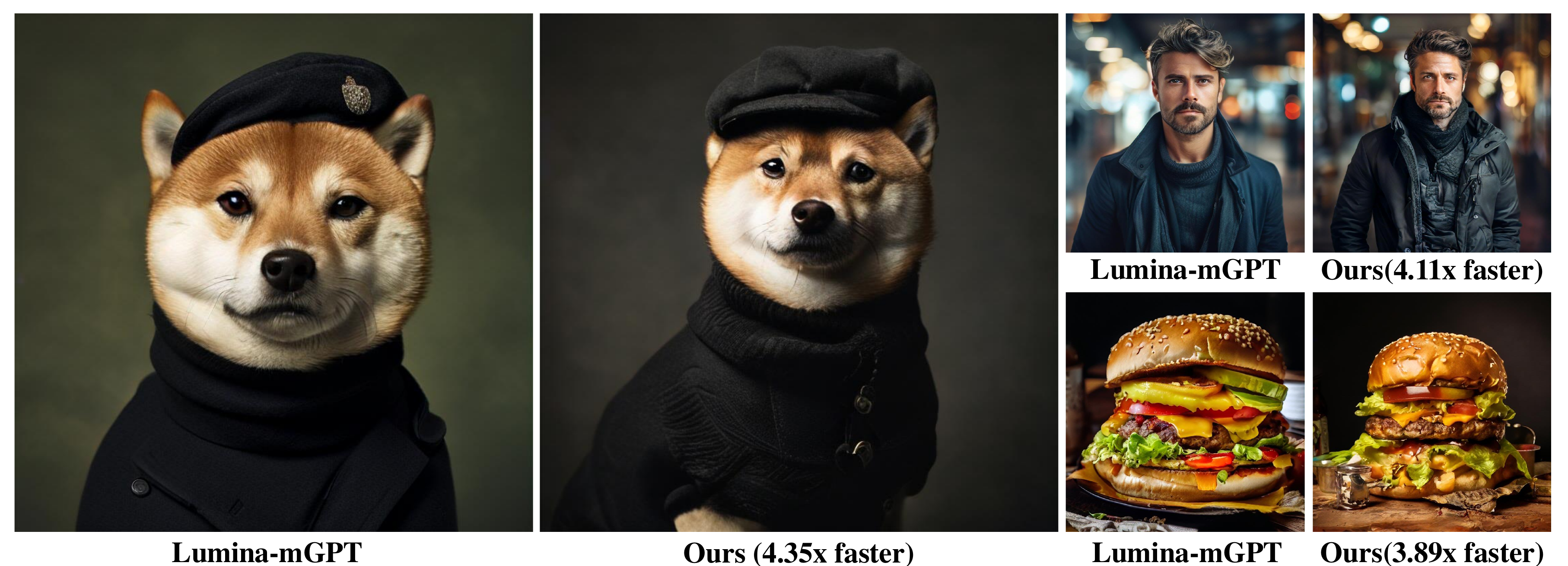} 
    \captionof{figure}{Our \textbf{\textit{Grouped Speculative Decoding}} (GSD) accelerates autoregressive image generation up to 4.3x without compromising quality. }
\end{center}%
}]

\begin{abstract}
Recently, autoregressive (AR) image models have demonstrated remarkable generative capabilities, positioning themselves as a compelling alternative to diffusion models. However, their sequential nature leads to long inference times, limiting their practical scalability. In this work, we introduce Grouped Speculative Decoding (GSD), a novel, training-free acceleration method for AR image models. While recent studies have explored Speculative Decoding (SD) as a means to speed up AR image generation, existing approaches either provide only modest acceleration or require additional training. Our in-depth analysis reveals a fundamental difference between language and image tokens: image tokens exhibit inherent redundancy and diversity, meaning multiple tokens can convey valid semantics. However, traditional SD methods are designed to accept only a single most-likely token, which fails to leverage this difference, leading to excessive false-negative rejections. To address this, we propose a new SD strategy that evaluates clusters of visually valid tokens rather than relying on a single target token. Additionally, we observe that static clustering based on embedding distance is ineffective, which motivates our dynamic GSD approach. Extensive experiments show that GSD accelerates AR image models by an average of 3.7× while preserving image quality—all without requiring any additional training. The source code is avaliable at \hyperlink{here}{https://github.com/junhyukso/GSD}.
\end{abstract}

\renewcommand\thefootnote{}\footnote{This paper was accepted to the International Conference on Computer Vision (ICCV) 2025.}
\addtocounter{footnote}{-1}
    
\section{Introduction}

\label{sec:intro}

Recent advancements in Large Language Models (LLMs) \cite{gpt4, llama, qwen} have demonstrated the remarkable ability of autoregressive (AR) models to capture highly complex distributions. This success has sparked widespread efforts to extend AR models to various domains \cite{valle, cosmos, protein}. In particular, AR image models \cite{llamagen, luminangpt, chameleon} have achieved impressive generative performance, establishing themselves as a strong alternative to diffusion models \cite{ddpm, ddim, dit}. Compared to diffusion-based approaches, AR image models offer several key advantages, including flexible resolution 
and seamless support for multimodal tasks.

Despite these advantages, AR image generation faces a major challenge: \emph{sequential token-by-token generation}. Generating a high-resolution image can require an AR model to produce \textbf{thousands} of tokens in sequence, whereas diffusion models typically need only \textbf{dozens} \cite{ddim}, despite their own limitations due to iterative refinement. This strict sequential dependency imposes a heavy computational burden, leading to significant latency that hinders the practical use of AR models in latency-critical scenarios. Furthermore, unlike text generation—where users can start reading as tokens appear—image generation demands that all pixels be rendered before any meaningful output is visible. This \textbf{all-or-nothing nature} of AR image generation results in a frustrating user experience, making it less suitable for interactive applications.

To tackle this challenge, various acceleration techniques have been explored \cite{wang2024continuous, wang2024parallelized, chen2024collaborative}. One promising approach is Speculative Decoding (SD) based techniques~\cite{SJD, LANTERN}, originally developed and validated in the context of LLMs~\cite{SD}. In its representative implementation, SD \cite{SD} leverages a lightweight draft model to propose token predictions rapidly, which are then verified in parallel by a larger base model. This allows for the simultaneous generation of multiple tokens, significantly boosting the AR generation speed of language tokens. However, when applied to AR image generation, existing SD-based methods have achieved only modest speed-ups \cite{SJD} or required additional training for the draft model \cite{LANTERN}, limiting their practicality.

In this work, we conduct a thorough investigation into \textbf{why SD underperforms in AR image generation}. We identify that \textbf{the inherent redundancy and diversity among visual tokens} results in relatively \textbf{low and uniformly distributed next-token probabilities} during decoding, which significantly reduces SD's acceptance rate.  
Building on this insight, we propose \textbf{Grouped Speculative Decoding (GSD)}—a novel technique that performs SD at the level of semantically valid token groups rather than focusing on a single most-likely token. We theoretically and empirically demonstrate that GSD boosts the acceptance rate with a simple yet effective modification to the acceptance criterion.  Additionally, we show that static clustering methods are suboptimal, leading us to introduce \textbf{Dynamic GSD}, which dynamically adjusts token grouping based on contextual information. Extensive experiments show that \textbf{GSD achieves an average 3.8× speedup} while maintaining high image quality, making it a practical and effective solution for accelerating AR image generation.

\section{Preliminaries }
\label{sec:prelim}

\subsection{Autoregressive Image model}

The modern AR image models \cite{llamagen, luminangpt,chameleon} typically consist of two components: Vector Quantizer (VQ)\cite{vqvae, vqgan} and an Autoregressive Transformer \cite{gpt3}. The VQ discretizes continuous images patches into discrete tokens through three main elements: an Encoder \(\mathcal{E}\), a Decoder \(\mathcal{D}\), and a Codebook \( C = \{c_1, c_2, \dots, c_n\} \), where each code \( c_i \in \mathbb{R}^d \).

Formally, the \textbf{encoder} maps an input image \( X \in \mathbb{R}^{H \times W \times C} \) to a latent representation \( x \in \mathbb{R}^{h \times w \times d} \), where \( h \) and \( w \) denote the spatial dimensions of the latent space, and \( d \) is the feature dimension. Each of the \( d\) dimensional feature vectors in this latent representation is then quantized by mapping it to the nearest code \( c \) in the \textbf{codebook}, resulting in the quantized representation \( x_q \). Finally, the \textbf{decoder} reconstructs the original RGB image from these quantized codes. The \textbf{encoder, codebook, and decoder} are jointly optimized by minimizing the reconstruction loss. This entire process can be summarized as follows:

\vspace{-0.3cm}
\[
\underbrace{X}_{\mathbb{R}^{H \times W \times C}} \xrightarrow{\mathcal{E}(X)} \underbrace{x \xrightarrow{\text{nearest}} x_q}_{\mathbb{R}^{h \times w \times d}} \xrightarrow{\mathcal{D}(x_q)} \underbrace{\hat{X}}_{\mathbb{R}^{H \times W \times C}}
\]

Using this codebook \( C \), the AR Image Model is trained to predict the probability distribution of the next token, just like text-based transformer models \cite{gpt3}. Recent studies \cite{chameleon, luminangpt} have explored a unified generation of both image and text tokens within a single vocabulary, allowing for a more efficient serving system under a unified architecture.

\begin{algorithm}[ht]
\caption{Speculative Decoding~\cite{SD}}
\begin{algorithmic}[1]
\Require Draft Length $L$, Maximum Length $N$, Draft model $q_\theta $, Target model $p_\theta $, Initial context $X_{0:n_0 }$
\State $n \gets n_{0}$
\While{$n < N$}
    \For{$j = 0$ to $L$} \Comment{AR Draft}
        \State $q_{j} \gets q(\cdot \mid [ X_{0:n}, \hat{X}_{0 : j}])$, $\hat{X}_{j} \sim q_{j}(\cdot)$  
    \EndFor
    \State \textbf{parallel for} $j = 0$ to $L$ \Comment{Parallel Verify}
        \State \quad $p_{j} \gets p(\cdot \mid [X_{0:n}, \hat{X}_{0 : j}])$
    \State \textbf{end for}
    \State $(\hat{X}_{0:k},\, k) \gets \texttt{VERIFY}(\hat{X}_{0:L}, p_{0:L}, q_{0:L})$
    \State $X_{n:n+k-1} \gets \hat{X}_{0:k}$, $n \gets n+k$ \Comment{Accept}
\EndWhile
\State \Return $X$
\end{algorithmic}
\label{alg:sd}
\end{algorithm}

\begin{algorithm}
\caption{\texttt{VERIFY}$(X,p,q)$}
\begin{algorithmic}[1]
\Require Draft $\hat{X}_{0:L}$, Verifier : $p_{0:L}(\cdot)$ , Drafter : $q_{0:L}(\cdot)$
\For{$k = 0$ to $L$}
    \If {$ \text{not\ }r \sim \mathcal{U}[0,1] \le \min \left( 1, \frac{p_k(\hat{X}_k)}{q_k(\hat{X}_k)} \right)$}
        \State  $x \sim [p_k - q_k]_+$ , $\hat{X}_k \gets x$, \textbf{break}.
    \EndIf
\EndFor
\State \Return $\hat{X}_{0:k}, k$
\end{algorithmic}
\label{alg:verify}
\end{algorithm}

\begin{figure*}[ht!]
    \centering
    \includegraphics[width=\textwidth]{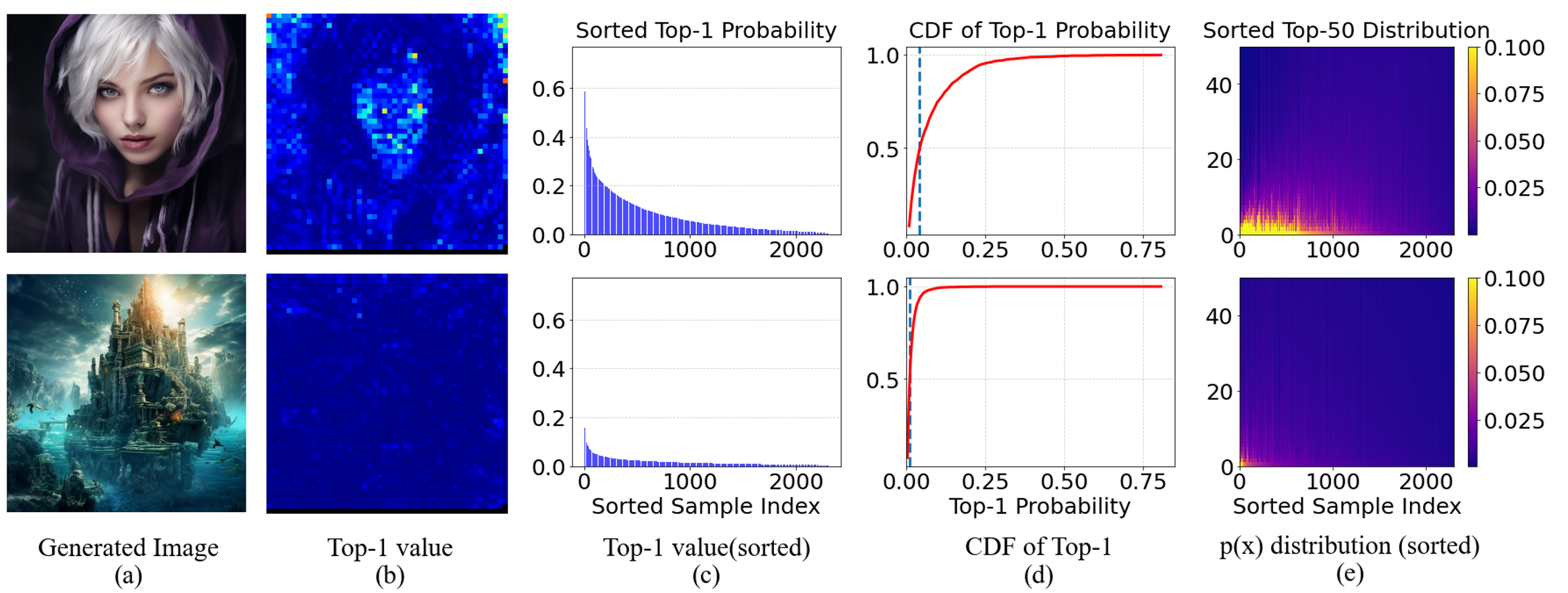} 
    \caption{Each patch's p(x) visualization. We used Lumina mGPT with a standard decoding setting ($\tau=1, K=2000$)}
    \label{fig:p_distribution}
\end{figure*}

\subsection{Speculative Decoding }
To accelerate AR decoding, SD \cite{SD} generates multiple tokens simultaneously by leveraging two distinct distributions: a precise expert (base) model \( p_\theta(\cdot) \) and a more efficient draft model \( q_\theta(\cdot) \). The process begins with the draft model \( q(\cdot) \) autoregressively generating \( L \) tokens. The expert model \( p(\cdot) \) then evaluates the exact probabilities of these tokens in parallel. Each token is sequentially accepted or rejected based on an acceptance probability of \( \min\left(1, \frac{p(x)}{q(x)}\right) \). If a token is rejected, SD resamples from an adjusted distribution \( [p - q]_+ \), where \( [\cdot]_+ \) represents \( \text{norm}(\max(0, \cdot)) \). After this step, a new batch of \( L \) tokens is generated by \( q(\cdot) \), and the process repeats. The complete procedure is outlined in Algorithm \ref{alg:sd} and Algorithm \ref{alg:verify}. Notably, this approach ensures that the base distribution \( p(\cdot) \) is precisely recovered while sampling from the draft distribution \( q(\cdot) \), enabling significant acceleration of AR models without compromising output quality.

\subsection{Speculative Jacobi Decoding}

While the original SD~\cite{SD} uses a separate, lightweight model to generate the draft, this is not a strict requirement. Since the correctness of the output distribution is ensured regardless of the choice of \( q \), other methods with more efficient output estimation can also be used for drafting. From this perspective, Speculative Jacobi Decoding (SJD) \cite{SJD} introduces an alternative approach by leveraging early predictions from a Jacobi iteration \cite{jacobi} as the draft distribution. Specifically, SJD starts by randomly initializing \( L \) tokens \( x_{0:L} \) and performs a parallel forward pass—similar to the parallel computation in SD—to estimate the output distribution \( p(x)_{0:L} \). The tokens are then sequentially verified using the same verification function, \( \text{VERIFY}(\cdot) \). A detailed description of the algorithm is provided in Algorithm~\ref{alg:sjd}.

This approach is highly efficient because verification and drafting occur simultaneously within a single parallel inference step, eliminating the need for autoregressive inference of \( q \) or training a dedicated draft model. Furthermore, since rejected tokens are reused, they gradually converge toward their correct values over iterations, leading to a higher acceptance rate. As a result, SJD achieves up to a 2.1× speedup across various text-to-image AR models. Due to these advantages, \textbf{we use SJD as our baseline and introduce novel ideas on top of it throughout the rest of this paper.}


\section{Low Acceptance Rate of SD in Image AR}
\label{sec:low_acceptance}

Although SJD achieves a meaningful speedup, its acceptance rate in SD remains relatively low (around 40\%) compared to that in typical text SD (around 70\%) \cite{SD}. In this section, we provide an in-depth analysis of why the acceptance rate for SD in image AR is relatively low.

\subsection{Characteristics of Image AR Decoding}
\label{sec:image_ar_char}

The figure~\ref{fig:p_distribution} provides a visual representation of \textit{(a)} images generated by the AR model , \textit{(b)} the top-1 probability values of the corresponding patchs and \textit{(c)--(e)} further illustrate the sorted distribution of these top-1 probabilities across all tokens. As shown in the figure, the image AR model frequently assigns low top-1 score (below 5\%) to 50\%–95\% of tokens, depending on the prompt. These probabilities are nearly uniformly distributed, suggesting that the model considers multiple tokens as equally plausible next steps. We attribute this phenomenon to two key factors:

\begin{itemize}
    \item \textbf{Redundancy in visual tokens :} 
    Unlike discrete text tokens, visual tokens are derived through vector quantization~\cite{vqvae} from a continuous latent space. Although quantized, these visual tokens still contain significant redundancy in low-frequency components, differing primarily in subtle, high-frequency details that are often imperceptible to the human eyes~\cite{vqgancollapse}.

    \item \textbf{Diversity of image patches :} 
    Unlike text, which is constrained by syntax and grammar, images can accommodate multiple valid visual patterns. For example, in Figure~\ref{fig:p_distribution}.(b), the model assigns lower top-1 probabilities to highly variable regions, such as hair, which can differ significantly in shape and texture. In contrast, more structured regions, like faces, tend to have higher top-1 probabilities due to their relatively consistent features.
\end{itemize}

\begin{figure}[t]
    \centering
    \scalebox{0.9}{  
        \begin{minipage}{\linewidth}
            \centering
            
            \begin{minipage}{0.5\linewidth}
                \centering
                \vspace{-0.5cm}
                \includegraphics[width=\linewidth]{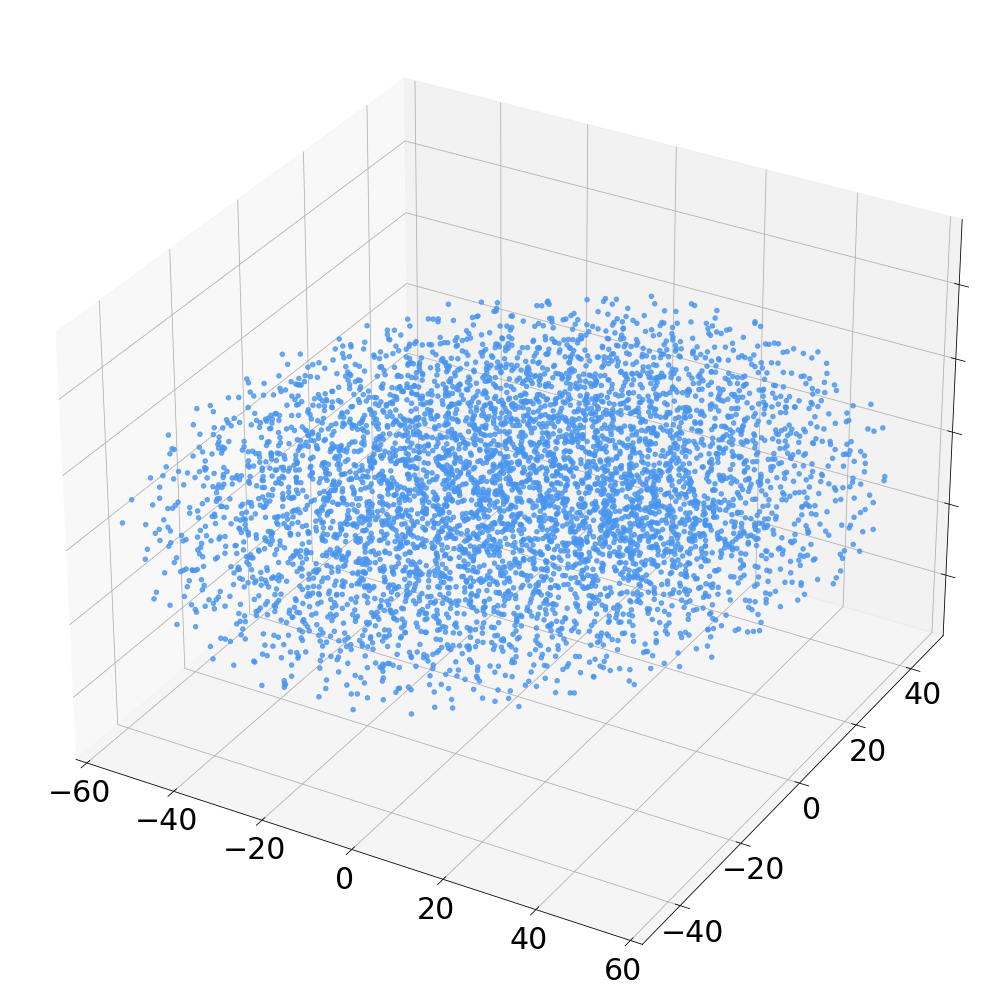} 
                \vspace{-0.5cm}
                \caption{t-SNE\cite{van2008visualizing} 3D visualization of the visual token embedding of \cite{luminangpt}}
                \label{fig:vqvae-tsne}
            \end{minipage}
            \hfill  
            \begin{minipage}{0.4\linewidth}
                \centering
                \includegraphics[width=\linewidth]{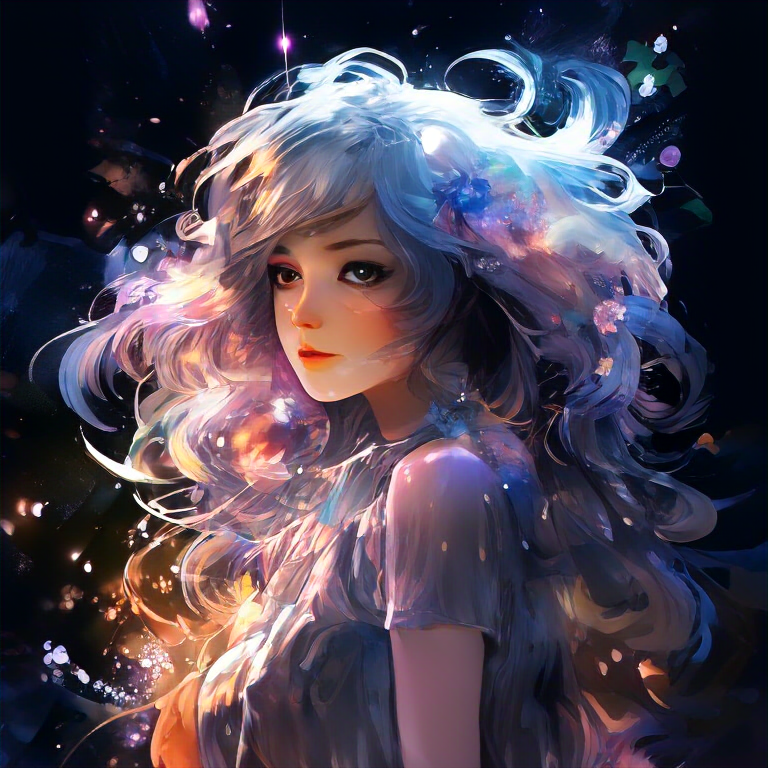}
                \caption{Result with 50\% random replacement during decoding, having $\text{CLIP\_score}=32.089$.}
                \label{fig:random_replace}
            \end{minipage}
        \end{minipage}
    }
\end{figure}

In Figure \ref{fig:random_replace}, we present an additional experiment where 50\% of the tokens are randomly replaced with their Top-100 candidates during decoding. As shown, the overall image quality remains largely unaffected, suggesting that the model indeed considers multiple valid next-step tokens and that substituting them has minimal impact. However, these findings raise a crucial question: Intuitively, if the model \( p(x) \) is indifferent to many possible token choices, the draft model shouldn’t need to be highly accurate, which should, in turn, lead to higher acceptance rates. \textbf{So why does speculative decoding still slow down in image AR?}

\subsection{Total Variation Analysis}
\label{sec:tv_analysis}

To explore this question, we start with following definition. From \cite{SD}, SD's expected acceptance rate $\alpha$ is defined as :
\vspace{-0.3cm}
\begin{equation}
     \textbf{Definition. }\alpha_{p,q} = \mathbb{E}_{x\sim q(x)}\left[\min\left(1, \frac{p(x)}{q(x)}\right)\right].
     \label{eq:defalpha}
\end{equation}
This directly leads to the following proposition:

\begin{tcolorbox}[colback=gray!10,colframe=gray!10,coltitle=black,title=\textbf{Proposition 1 (Acceptance and Total Variation)}] 
\emph{let $\alpha_{p,q}$ be acceptance rate defined on Eq.\ref{eq:defalpha}, and $TV(\cdot,\cdot)$ be Total Variation distance measure, then }

    $\alpha_{p,q} = 1-\frac{1}{2}\sum_x |p(x) - q(x)| \;=\; 1 - \mathrm{TV}(p, q).$
\end{tcolorbox}
\textit{Proof. See Appendix.} A similar analysis has been explored in previous works \cite{SD, SDtheory}. As shown, the expected acceptance rate for each token is determined by the absolute difference between the two probability distributions: the expert model \( p \) and the draft model \( q \).

\begin{figure}[t]
    \centering
    \begin{subfigure}[b]{0.22\linewidth}
        \centering
        \includegraphics[width=\linewidth]{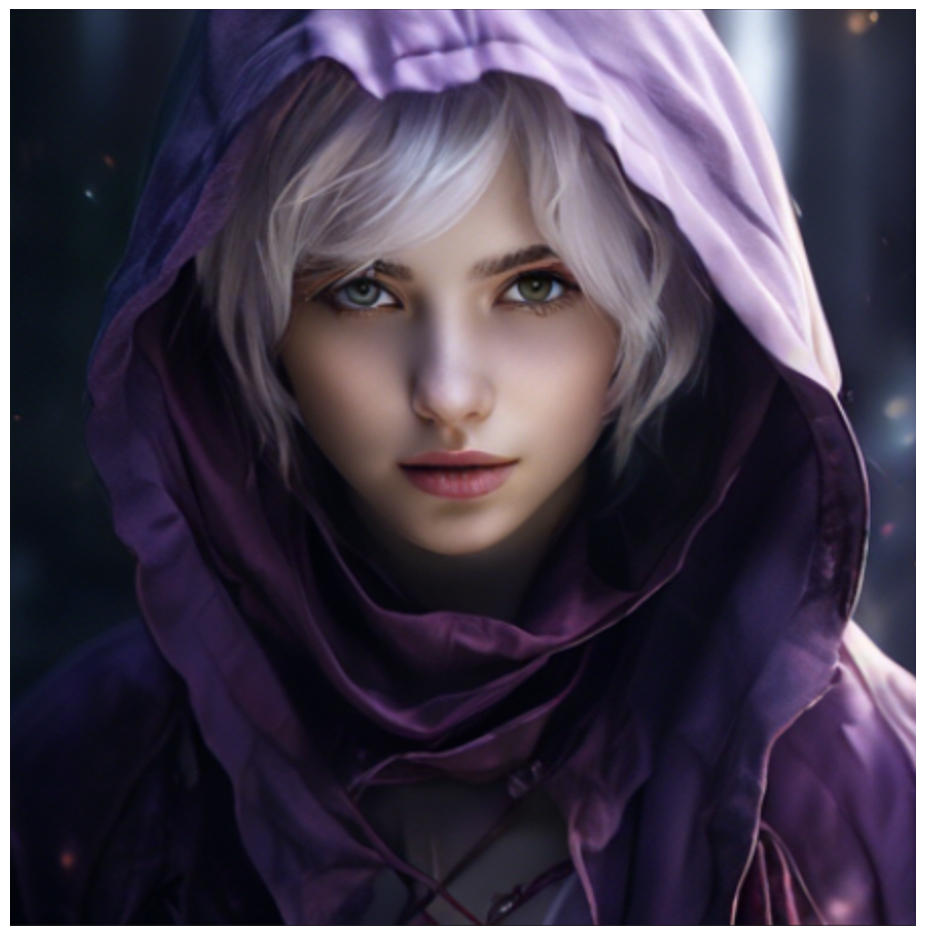}
        \caption{Image}
        \label{fig:genimg}
    \end{subfigure}
    \begin{subfigure}[b]{0.24\linewidth}
        \centering
        \includegraphics[width=\linewidth]{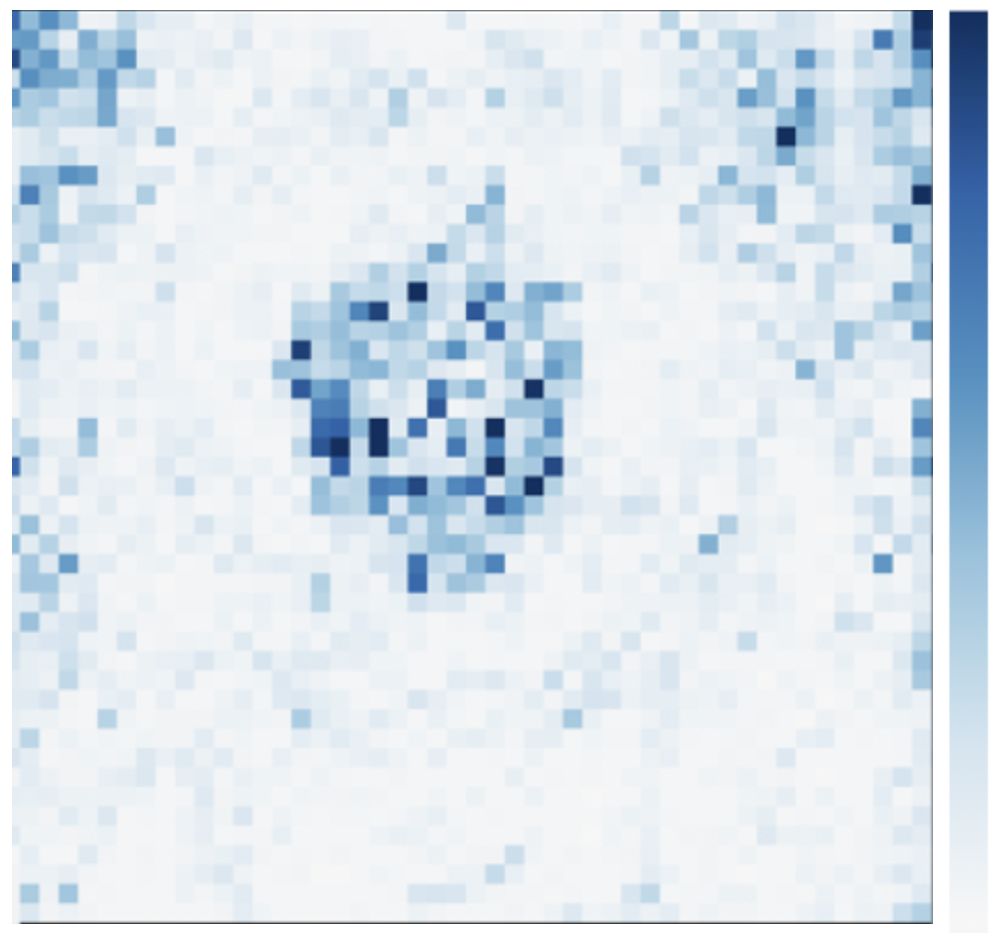}
        \caption{$p(x)$ Top-1 }
        \label{fig:ptop1}
    \end{subfigure}
    \begin{subfigure}[b]{0.24\linewidth}
        \centering
        \includegraphics[width=\linewidth]{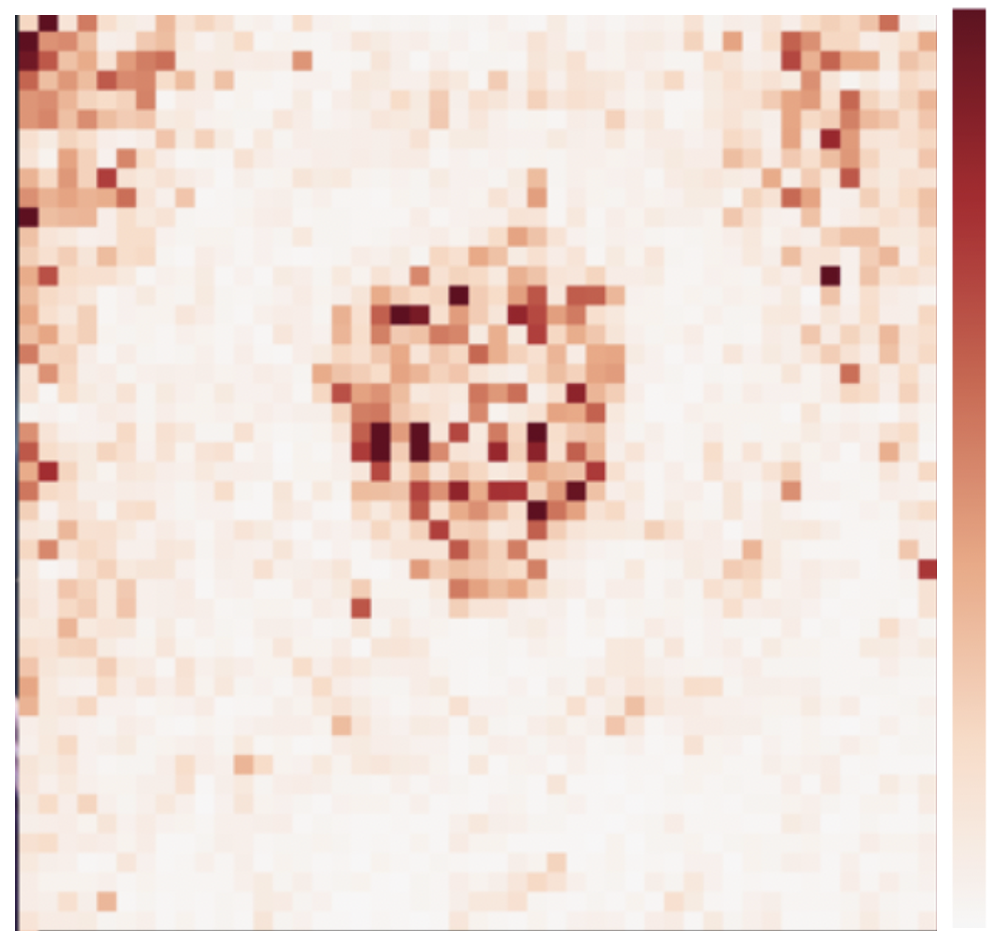}
        \caption{$q(x)$ Top-1  }
        \label{fig:qtop1}
    \end{subfigure}
    \begin{subfigure}[b]{0.26\linewidth}
        \centering
        \includegraphics[width=\linewidth]{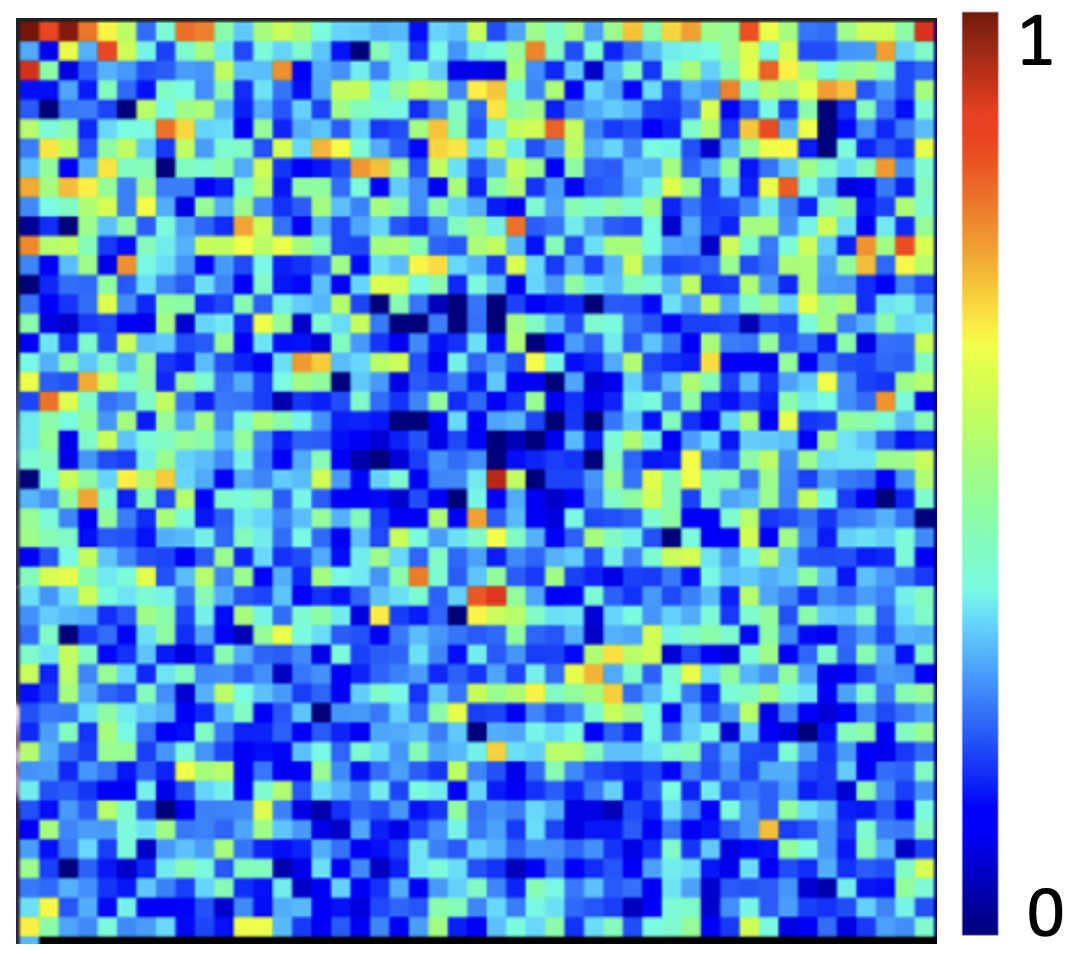}
        \caption{$TV(p,q)$}
        \label{fig:tv}
    \end{subfigure}

    \caption{Visualization of image, $p(x)$ Top-1, $q(x)$ Top-1 and $TV(p,q)$. While both $p,q$ have small Top-1, their $TV$ is high.}
    \label{fig:pt_qt_distributions}
\end{figure}

\begin{figure}[t]
    \centering
    \includegraphics[width=0.98\linewidth]{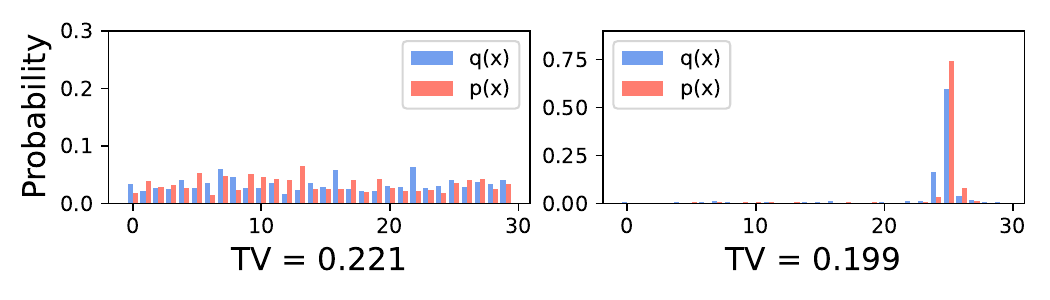}
    \vspace{-0.5cm}
    \caption{A toy example of the accumulation problem. While the left tends to have an almost uniform distribution, its total variation (TV) is larger than that of the right. This problem becomes more pronounced when the support is larger.}
    \label{fig:toy_example}
\end{figure}

In Figure~\ref{fig:pt_qt_distributions}, we present \textit{(a)} a generated image, \textit{(b)} the top-1 probability of \( p(x) \), \textit{(c)} the top-1 probability of \( q(x) \), and \textit{(d)} the total variation \( \mathrm{TV}(p, q) \) for each patch. Notably, even when both distributions \( p(x) \) and \( q(x) \) agree that multiple tokens are plausible—indicated by low top-1 probabilities—the total variation between them remains high. Conversely, regions with higher confidence (e.g., faces) exhibit relatively low TV values. These results suggest that although many tokens are valid as the next choice, \textbf{the accumulation of subtle differences between distributions \( p \) and \( q \) significantly increases total variation}, ultimately lowering the acceptance rate. To illustrate this concept more clearly, we provide a toy example in Figure~\ref{fig:toy_example}. In practice, this effect is further amplified by the large vocabulary size—approximately 20,000 tokens—used in recent AR image models \cite{chameleon, luminangpt}.

\section{Method: Grouped Speculative Decoding}
\label{sec:method_gsd}
To overcome this issue, we introduce a novel approach called \textit{Grouped Speculative Decoding} (GSD), which bases acceptance decisions on \textbf{semantically meaningful tokens} rather than solely selecting the most likely token.

Specifically, let \( \mathcal{X} \) represent the token vocabulary, and let \( \textrm{p}(x) \) and \( \textrm{q}(x) \) be the probability mass functions (p.m.f.) of the expert and draft models, respectively, defined over \( \mathcal{X} \). We then partition \( \mathcal{X} \) into \textit{disjoint clusters}:
\begin{equation}
\mathcal{C} \;=\; \{C_1,\;C_2,\;\dots,\;C_K\},
\label{eq:cluster}
\end{equation}
where $C_i \cap C_j = \varnothing$ for $i \neq j$ and $\bigcup_{i=1}^K C_i = \mathcal{X}$. For each cluster $C_i$, we define the grouped probability mass as:
\begin{equation}
  \textrm{p}'(C_i) \;=\; \sum_{x \in C_i} \textrm{p}(x), 
  \quad
  \textrm{q}'(C_i) \;=\; \sum_{x \in C_i} \textrm{q}(x).
  \label{eq:group}
\end{equation}

Consequently, \( \textrm{p}' \) and \( \textrm{q}' \) are also p.m.f. defined over \( \mathcal{C} \), satisfying \( \sum_{C_i} \textrm{p}'(C_i) = 1 \), with a similar condition holding for \( \textrm{q}' \). We denote by \( C(\cdot) \) the mapping \( \mathcal{X} \to \mathcal{C} \) that assigns each token \( x \) to its corresponding cluster \( C_i \). The specific implementation of \( C(\cdot) \) will be detailed in the next section.

In GSD, we use the identical SD verification procedure but decide to accept a token $x$ if its cluster $C(x)$ satisfies:
\begin{equation}
    \min\!\Bigl(1, \frac{\textrm{p}'\bigl(C(x)\bigr)}{\textrm{q}'\bigl(C(x)\bigr)}\Bigr) \;\ge\; r, \quad r \sim \mathcal{U}[0,1].
    \label{eq:gsd_criterion}
\end{equation}
If rejected, we follow the same resampling strategy as in standard SD. Intuitively, by summing individual masses within each cluster, we obtain a more coarse-grained probability distribution, smoothing out subtle differences between $\textrm{p}$ and $\textrm{q}$. This reduces $\mathrm{TV}(\textrm{p}', \textrm{q}')$ and thereby improves the acceptance rate. We formally state this in the theorem below:

\begin{tcolorbox}[colback=gray!10,colframe=gray!10,coltitle=black,title=\textbf{Theorem 1. Lower Bound on GSD Accept Rate}]
\emph{Let $\textrm{p}, \textrm{q}$ be p.m.f defined over $\mathcal{X}$, let $\mathcal{C}$ is disjoint clstuer defined by Eq. \eqref{eq:cluster} and let $\textrm{p}', \textrm{q}'$ be the corresponding grouped p.m.f defined by Eq. \eqref{eq:group}. Then for any choice of $(\textrm{p},\textrm{q},\mathcal{C})$, the acceptance rate of GSD is bounded below by the acceptance rate of standard SD, meaning that $\alpha_{GSD} \ge \alpha_{SD}$}
\end{tcolorbox}

\noindent\textit{Proof Sketch.}
From Proposition 1, to show $\alpha_{GSD}>\alpha_{SD}$, we sufficient to show $\mathrm{TV}(p', q') \le \mathrm{TV}(p, q)$. By Eq. \eqref{eq:group}, the total variation of $p',q'$ and $p,q$ can be expaned as :
\begin{align*}
   \mathrm{TV}(p', q') &= \frac{1}{2} \sum_{C_i} \bigl| p'(C_i) - q'(C_i) \bigr| \\ 
   &= \frac{1}{2} \sum_{C_i} \Bigl| \sum_{x\in C_i} \bigl( p(x) - q(x) \bigr) \Bigr|. \\
   \mathrm{TV}(p, q) &= \frac{1}{2} \sum_{x\in\mathcal{X}} \bigl| p(x) - q(x) \bigr|  \\
   &= \frac{1}{2} \sum_{C_i} \sum_{x\in C_i} \Bigl| p(x) - q(x) \Bigr|.
\end{align*}

\noindent Because of the triangle inequality, for any C we have,
\[
   \Bigl|\sum_{x\in C_i} \bigl(p(x) - q(x)\bigr)\Bigr|
   \;\le\; \sum_{x\in C_i} \bigl|p(x) - q(x)\bigr|.
\]
Summing it over all $C$, we have $\mathrm{TV}(p', q') \le \mathrm{TV}(p,q)$ $\qedsymbol$

\begin{figure}[t]
    \centering
    \includegraphics[width=\columnwidth]{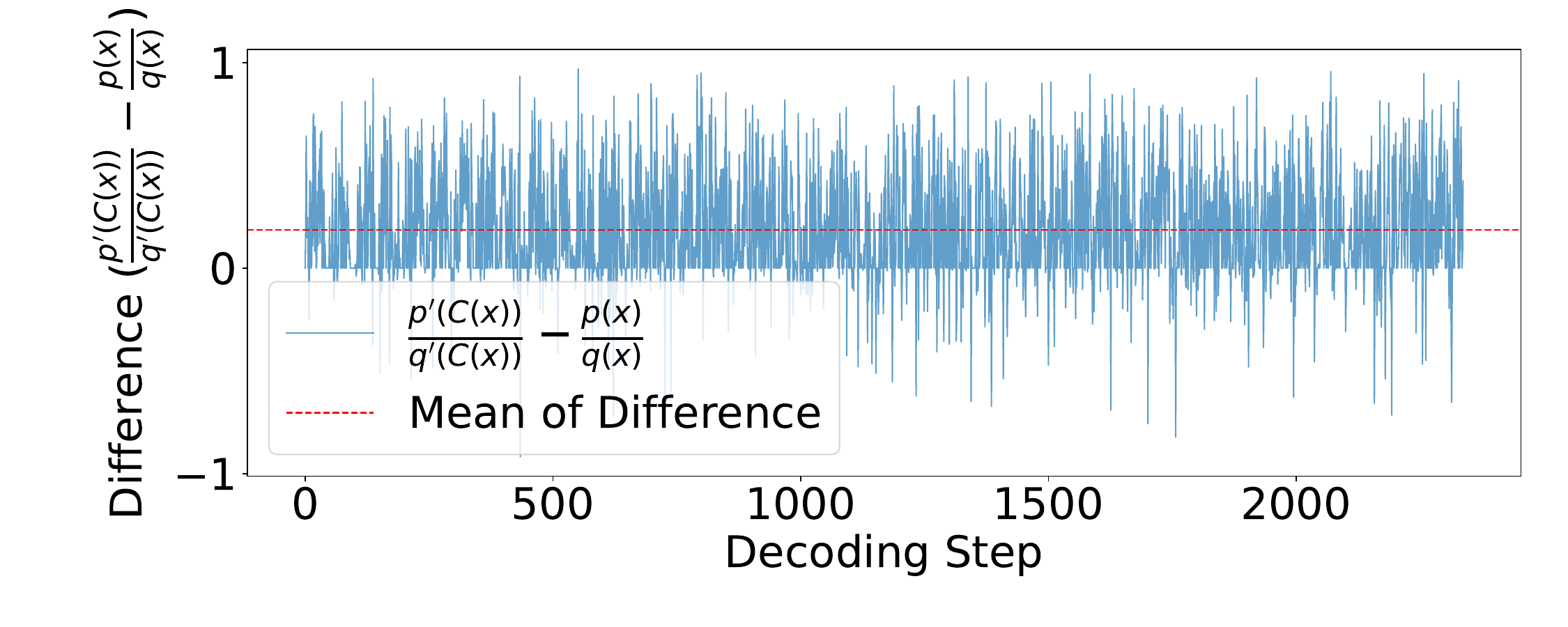} 
    \caption{Diffrence of acceptance proability ($ \frac{p'(C(x))}{q'(C(x))} - \frac{p(x)}{q(x)}$) during image AR decoding. }
    \vspace{-0.5cm}
    \label{fig:diffrence}
\end{figure}

As shown,  while $\frac{p'(C(x))}{q'(C(x))} \ge \frac{p(x)}{q(x)}$ may not hold for every individual token $x$,  its expected rate is still higher. To support our theoretical anlysis, we depicts the probability difference during image AR decoding in Fig. \ref{fig:diffrence}. As shown, most tokens have greater acceptance probability in cluster level and mean difference is actually positive value, validating our theoretical result.

\subsection{Context-aware Dynamic Clustering}

The key takeaway from Theorem~1 is that GSD improves or at least preserves decoding speed regardless of how clusters are formed. However, designing an \emph{effective} clustering strategy remains critical for maintaining output quality.

\begin{figure}[t!]
    \centering
    \begin{subfigure}[b]{0.3\columnwidth}  
        \centering
        \includegraphics[width=\linewidth]{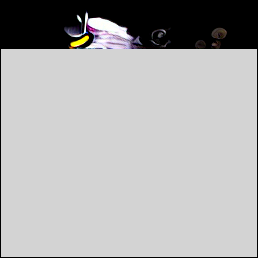}
        \caption{20\%}
    \end{subfigure}
    \begin{subfigure}[b]{0.3\columnwidth}
        \centering
        \includegraphics[width=\linewidth]{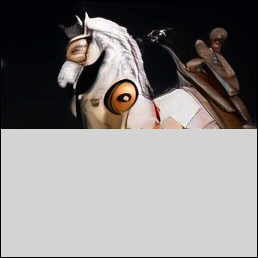}
        \caption{50\%}
    \end{subfigure}
    \begin{subfigure}[b]{0.3\columnwidth}
        \centering
        \includegraphics[width=\linewidth]{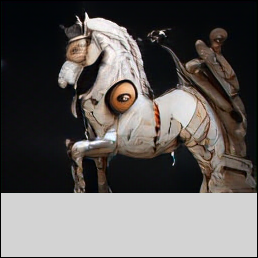}
        \caption{80\%}
    \end{subfigure}
    \caption{Visualization of decoded results during generation with different numbers of rows. Even though tokens are identical, the resulting RGB images differ in details and color, indicating that accurate clustering should consider the context of nearby tokens.
}
\vspace{-0.3cm}
\label{fig:decoded_row}
\end{figure}

\textit{How can we form effective clusters?} A straightforward approach is to cluster tokens based on pairwise distances among their corresponding codebook embeddings in the embedding space. However, we observed that this strategy frequently underperforms, as shown in Table \ref{tab:cluster}. We attribute this primarily to two reasons:

\begin{itemize}
    \item \textbf{Uniformity of the codebook embedding space :} Figure~\ref{fig:vqvae-tsne} shows a t-SNE \cite{van2008visualizing} 3D visualization of the visual token embeddings obtained from \cite{luminangpt}. As illustrated, embeddings are distributed nearly uniformly, making it difficult to construct semantically coherent clusters. This phenomenon becomes more pronounced as the codebook have higher utilization \cite{vqgancollapse}.

    \item \textbf{Impact of token context in image :} Figure~\ref{fig:decoded_row} visualizes decoded results during generation using varying numbers of rows. Although tokens remain identical , their decoded RGB representations significantly differ, particularly in colors and fine details, highlighting the substantial influence of context. This effect becomes more pronounced when fewer tokens are present.
\end{itemize}

\noindent These observations offer an important insight: the raw embedding space does not fully represent a semantic manifold for image tokens. Instead, semantic meanings emerge when considering the token context, as thousands of tokens jointly input to the decoder.

\vspace{-0.3cm}
\paragraph{Context-aware Dynamic GSD.} Inspired by this insight, we propose leveraging the token probability $p(x)$ itself as a dynamic \textit{measure} of token similarity. This choice is motivated by the fact that the model’s predicted probability inherently reflects its perception of token similarity within the current decoding context. 

Specifically, at each decoding step $t$, tokens are sorted according to their probability values, and clusters are formed by grouping the number of $G$ nearest tokens. To avoid grouping tokens with significantly different semantics, we further exclude tokens whose embedding distances or probability differences exceed predefined thresholds $d$ and $\delta$. In Algorithm. \ref{alg:gsdverify}, we show psuedo-code of our Dynamics GSD verification. The complete detailed algorithm for our final method is provided in the Algorithm. \ref{alg:gsd}.

Importantly, as established by Theorem~1, even if the clustering function dynamically changes at each decoding step $t$, GSD still guarantees a higher acceptance rate than standard SD. This dynamic clustering not only improves the final image quality but also accelerates the decoding speed, as we will demonstrate in the next experimental section.

\begin{algorithm}
\caption{\texttt{VERIFY\_GSD}$(X,p,q,G)$}
\begin{algorithmic}[1]
\Require Draft $\hat{X}_{0:L}$, Verifier : $p_{0:L}(\cdot)$ , Drafter : $q_{0:L}(\cdot)$, Group size $G$, Embedding distance matrix $M_d$, thresholds $d,\delta$
\For{$k = 0$ to $L$} 
  \State $\rhd$ Dynamic Clustering with $p(\cdot)$
    \State $p\_sort_{vals},p\_sort_{idx} \gets \texttt{sort}(p_k)$
    \State $idx \gets \texttt{find-idx}(p\_sort_{vals}, p_k(\hat{X}_k))$
    \State $C_{idxs} \gets p\_sort_{idx}[idx -G//2 : idx + G//2]$
    \State $Cvals \gets p_k[C_{idxs}]$
    \\
    \State $\rhd$ Filter Outliers
    \For{$cv,ci$ in $[Cvals, C_{idxs}]$}
        \State $\text{if } |cv -p_k(\hat{X}_k)| > \delta \text{ then  $C_{idxs}$.pop($ci$)} $
        \State $\text{if }M_d[\hat{X}_k, ci] > $d$ \text{ then  $C_{idxs}$.pop($ci$)} $
    \EndFor
    \\

    \State$\rhd$ Verification with Cluster Probability
    \State $p'_C \gets \texttt{sum}(p_k[C_{idxs}])$
    \State $q'_C \gets \texttt{sum}(q_k[C_{idxs}])$
    \If {$ \text{not\ }r \sim \mathcal{U}[0,1] \leq \min \left( 1, \frac{p'_C}{q'_C} \right)$}
        \State  $x \sim [p_k - q_k]_+$ , $\hat{X}_k \gets x$, \textbf{break}.
    \EndIf
\EndFor
\State \Return $\hat{X}_{0:k}, k$
\end{algorithmic}
\label{alg:gsdverify}
\end{algorithm}

\begin{figure*}[t!]
    \centering
    \includegraphics[width=0.99\textwidth]{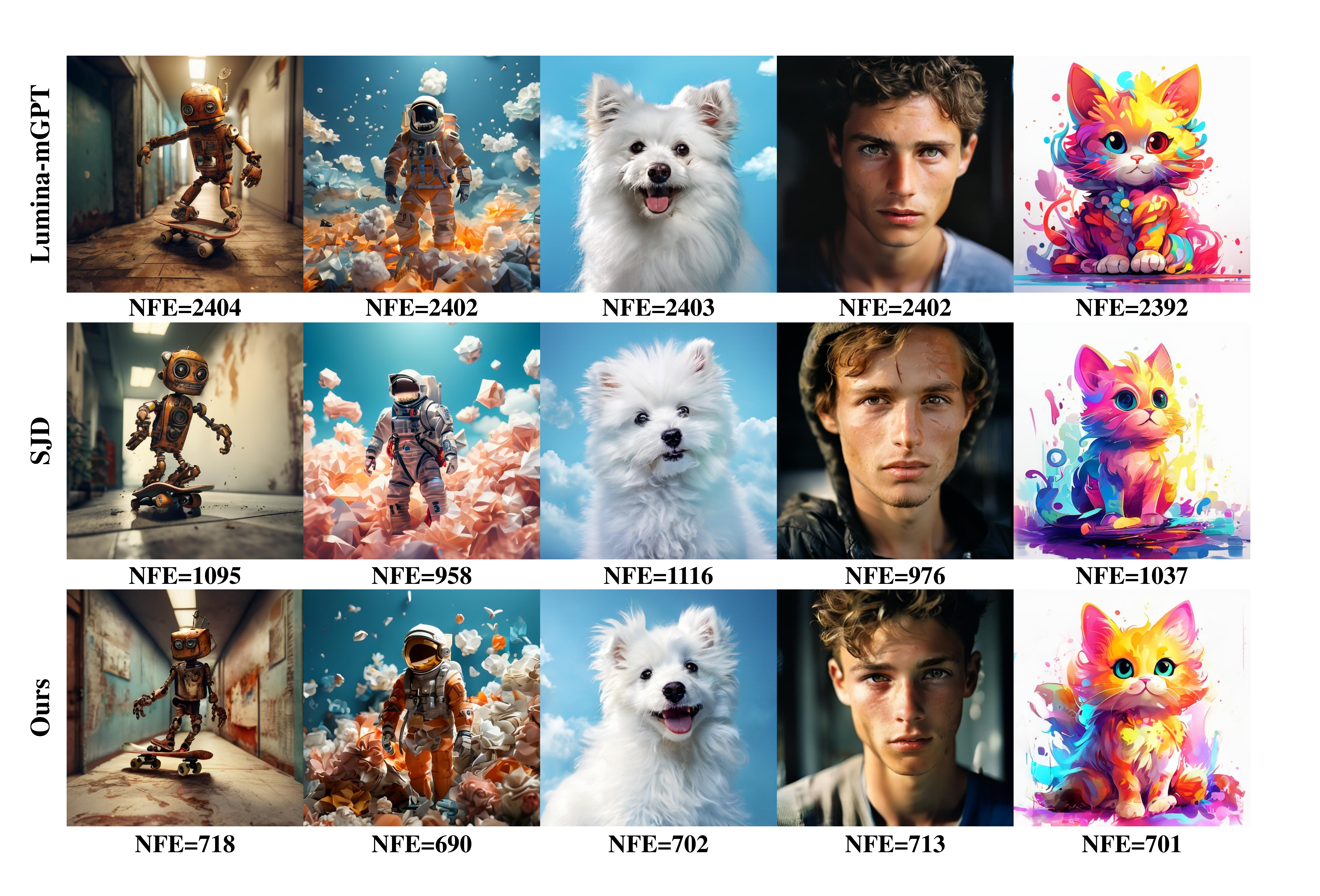} 
    \vspace{-0.5cm}
    \caption{Qualitative experiment. Our GSD shows on average 3.6x NFE acceleration while maintaining image quality}
    \label{fig:qual}
    \vspace{-0.3cm}
\end{figure*}

\section{Experiments}

\begin{table*}[t!]
    \centering
    \caption{Quantitative evaluation on the MS-COCO 2017 and Parti-prompt.}
    \begin{adjustbox}{max width=\linewidth}
    \begin{tabular}{l c c c c c c}
        \toprule
        \textbf{Configuration} & \textbf{Latency (↓)} & \textbf{NFE (↓)} & \multicolumn{2}{c}{\textbf{Acceleration} (↑)} & \textbf{FID (↓)} & \textbf{CLIP-Score (↑)} \\
        \cmidrule(lr){4-5}
        & & & \textbf{Latency} & \textbf{NFE} & & \\
        \midrule
        \multicolumn{7}{l}{\textbf{Parti-prompt }} \\
        A \quad Lumina-mGPT \cite{luminangpt} & 112.29s & 2392 & 1.00× & 1.00× & -- & 32.091  \\
        B \quad Jacobi Decoding \cite{jacobi} & 116.31s & 2300.0 & 0.97x & 1.04× & -- & 32.091 \\
        B \quad SJD \cite{SJD} & 52.34s & 1035.3 & 2.15x & 2.31× & -- & 32.090 \\
        \midrule
        C \quad Amplify (k=2) & 37.48s & 705.13 & 3.00x & 3.39× & -- & 31.906 \\
        C \quad Amplify (k=3) & 31.31s & 586.23 & 3.59x & 4.08x & -- & 31.774 \\
        C \quad Addition ($\epsilon$=1e-1) & 40.01s & 755.05 & 2.81x & 3.17x & -- & 31.919 \\
        C \quad Addition ($\epsilon$=3e-1) & 24.93s & 644.57 & 4.50x & 3.71x & -- & 31.776 \\
        \midrule
        D \quad Ours (G=3) & 47.12s & \textbf{898.97} & \textbf{2.38x} & \textbf{2.66x} & -- & \textbf{32.125} \\
        D \quad Ours (G=50) & 24.13s & \textbf{636.75} & \textbf{4.65x} & \textbf{3.76x} & -- & \textbf{32.075} \\
        D \quad Ours (G=100) & 23.01s & \textbf{611.75} & \textbf{4.88x} & \textbf{3.91×} & -- & \textbf{31.975} \\
        \bottomrule
        \midrule
        \multicolumn{7}{l}{\textbf{MS-COCO 2017 }} \\
        A \quad Lumina-mGPT \cite{luminangpt} & 122.45s & 2379 & 1.00× & 1.00× & 30.79 & 31.308  \\
        B \quad Jacobi Decoding \cite{jacobi} & 121.16s & 2312 & 1.01x & 1.03x & 30.78 & 31.308 \\
        B \quad SJD \cite{SJD} & 55.26s & 1058.6 & 2.22x & 2.25× & 30.78 & 31.308 \\
        \midrule
        C \quad Amplify (k=2) & 35.98s & 738.96 & 3.40x & 3.22x & 34.79 & 31.18 \\
        C \quad Amplify (k=4) & 32.29s & 635.17 & 3.79x & 3.75x  & 40.05 & 30.99 \\
        C \quad Addition ($\epsilon$=1e-1) & 47.80s & 909.12 & 2.56x & 2.62x & 32.75 & 31.2707 \\
        C \quad Addition ($\epsilon$=3e-1) & 31.20s & 661.66 & 3.92x & 3.60x & 40.20 & 30.937 \\
        \midrule
        D \quad Ours (G=3) & 48.28s & \textbf{925.89} & \textbf{2.54x} & \textbf{2.57x} & \textbf{31.50} & \textbf{31.33} \\
        D \quad Ours (G=10) & 33.79s & \textbf{701.35} & \textbf{3.62x} & \textbf{3.39x} & \textbf{33.12} & \textbf{31.25} \\
        D \quad Ours (G=25) & 32.52s & \textbf{674.04} & \textbf{3.77x} & \textbf{3.53×} & \textbf{33.55} & \textbf{31.24} \\
        \bottomrule
        \midrule
    \end{tabular}
    \end{adjustbox}
    \label{tab:quant}
    \vspace{-0.3cm}
\end{table*}

For our experiments, we employed the SOTA text-to-image AR Image model, Lumina-mGPT \cite{luminangpt}. Specifically, we used the standard 7B model and conducted experiments at a resolution of 768×768. In all experiments, we followed the default settings: Top-K sampling with \(K=2000\) and temperature \(\tau=1\). For GSD, we set \(d=0.5\), \(\delta=0.15\), and \(L=16\). We did not conduct experiments with the Greedy setting (\(\tau=0\)), as it significantly degrades the quality of generated images \cite{SJD,luminangpt}. 
We used PyTorch 2.3 \cite{pytorch} on a high-performance server equipped with 8 RTX 3090 GPUs and an AMD EPYC 7402 processor.  

\vspace{-0.3cm}

\paragraph{Qualtitiatve Experiments}
We first present qualitative experiments comparing generated images from Vanilla AR, SJD~\cite{SJD}, and our proposed method. To thoroughly evaluate the robustness and effectiveness of our approach, we carefully select five prompts designed to capture a wide range of environmental conditions and generation challenges.

Specifically, in Figure \ref{fig:qual}, we illustrate the following cases: (1) Unrealistic images, showcasing the model's ability to generate novel and imaginative content in a zero-shot manner. (2) Realistic images, which require precise rendering of complex contextual details. (3) Human faces, a particularly challenging category where even minor discrepancies in facial features are easily noticeable. (4) Animation-style illustrations, highlighting the method's adaptability to stylized visual content. Detailed prompts corresponding to each case are listed in the appendix. As clearly demonstrated in the figure, our proposed method consistently achieves superior image quality across all prompt categories while delivering an impressive 3.5× speedup compared to Vanilla AR decoding.

\vspace{-0.3cm}
\paragraph{Quantitative Experiment} 
To quantitatively evaluate both the generation quality and speed of our method, we conducted experiments on two datasets: Parti-Prompt~\cite{yu2022scaling} and MS-COCO~\cite{mscoco}. For Parti-Prompt, we evaluated the generation quality using the CLIP Score~\cite{clip}, which measures the similarity of images to a given prompt, on a set of 1,600 text prompts. For MS-COCO, we used 5,000 prompts from the validation set and measured both the CLIP Score and the FID Score~\cite{mscoco}, where the latter was computed by comparing the quality of generated images against the validation set images. To evaluate generation speed, we measured the average latency required to generate a single image and the number of function evaluations (NFE), which indicates the number of model forward passes performed.

We benchmarked our method against three baselines: (A) \textit{Vanilla AR}, (B) Lossless methods, including Jacobi Decoding~\cite{jacobi} and SJD~\cite{SJD}, and (C) Naive Lossy SD, such as Amplify $\left(\frac{k \cdot p}{q}\right)$ and Addition $\left(\frac{p+\epsilon}{q}\right)$ introduced in \cite{SD,SDtheory}, along with (D) \textbf{Ours}. For (C) and (D), we applied a lossy acceptance criterion atop SJD \cite{SJD}.

As shown in Table \ref{tab:quant}, when $G=3$, our method achieves higher performance with fewer NFEs compared to both (A) Vanilla and (B) Lossless. This indicates that GSD's clustering effectively smooths minor fluctuations in the next-token distribution $p(x)$, positively influencing overall image quality by reducing the noise in model predictions. When we allow slight quality degradation ($G=50$), our approach achieves a 3.6× speedup compared to Vanilla and a 1.63× speedup over SJD, enabling highly efficient speculative decoding in image AR. Furthermore, compared to naive lossy methods (C), which drastically degrade image quality, our method gives large acceleration with minimal quality loss.
\begin{table}[t]
    \centering
    \begin{minipage}{0.8\linewidth}
        \centering
        \caption{Comparison of different clustering on Parti. }
        \begin{tabular}{l c c}
            \toprule
            \textbf{Method} & \textbf{NFE} & \textbf{CLIP Score} \\
            \midrule
            \textit{Baseline} (SJD \cite{SJD}) & 1035 & 32.090 \\
            \midrule
            (A) Embed distance & 913.80 & 32.081 \\
            (B) draft $q(x)$  & 685.31 & 31.791 \\
            (C) expert $p(x)$ & 636.24 & 32.056 \\
            \textbf{(D)} Ours & 636.75 & 32.075 \\
            \bottomrule
        \end{tabular}
        \label{tab:cluster}
    \end{minipage}
    \vspace{-0.5cm}
\end{table}

\subsection{Ablation Stuides}

\paragraph{Pareto-front Comparison} 

\begin{figure}[t]
    \centering
    \begin{subfigure}[b]{0.49\linewidth}
        \centering
        \includegraphics[width=\textwidth]{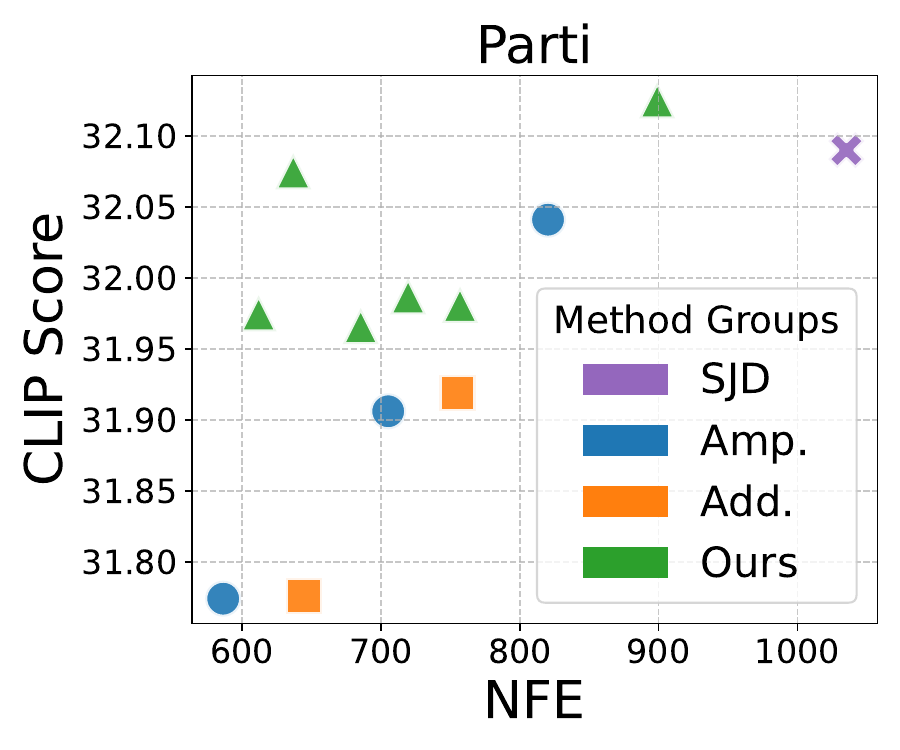}
    \end{subfigure}
    \hfill
    \begin{subfigure}[b]{0.49\linewidth}
        \centering
        \includegraphics[width=\textwidth]{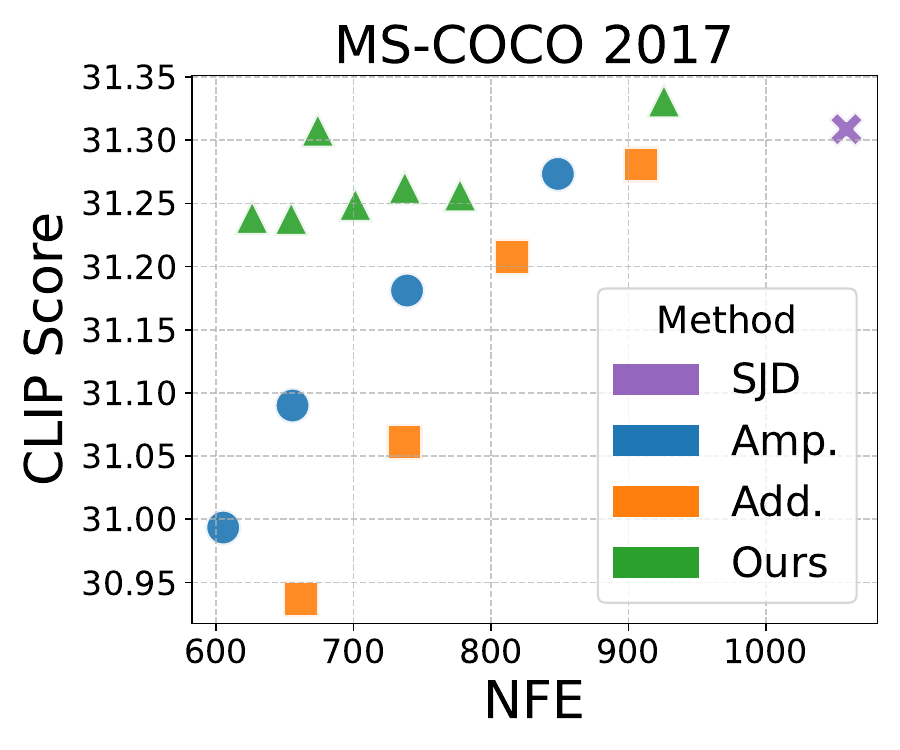}
    \end{subfigure}
    \caption{Pareto-front Comparision: NFE vs CLIP\_score}
    \vspace{-0.5cm}
    \label{fig:pareto}
\end{figure}

In Figure \ref{fig:pareto}, we expand on the experiments presented in Table \ref{tab:quant} by evaluating performance across a wider range of group sizes and visualizing these results using a Pareto front plot. As illustrated, our method achieves a superior Pareto front compared to naive lossy SD methods and notably also surpasses the lossless SJD by delivering higher performance with fewer NFEs. These findings indicate that GSD positively influences image quality by smoothing probabilities among similar tokens, thereby improving overall performance. Moreover, naive lossy methods suffer from drastic degradation due to bias or exploding behavior when $q(x)$ becomes small, as they only increase the numerator. In contrast, our method increases both $p$ and $q$ in an unbiased manner, effectively avoiding such issues and better preserving performance.

\paragraph{Effect of Cluster} 
In Table \ref{tab:cluster}, we evaluate the performance of GSD using three different clustering methods. As shown, (A) clustering solely based on static embedding distance fails to achieve speedup, indicating that probability is not strongly correlated with static embedding distance. (B) Clustering based on the draft $q(x)$ leads to speedup but results in a significant decline in the CLIP score, indicating that $q$ does not accurately capture token similarity when contexts are given. (C) When expert $p(x)$ is used for clustering, it successfully preserves the CLIP score while achieving acceleration. (D) Additionally, when filtering out tokens that exceed a certain threshold, performance improves.

\section{Related Works: Speculative Decoding}

After the pioneering work \cite{SD}, numerous studies have attempted to improve acceptance rates in SD. Mainstream approaches enhance acceptance while maintaining sampling exactness by proposing multiple drafts in batches \cite{spectr, SDtheory} or structuring them as trees \cite{eagle2,medusa}. Recently, a few studies \cite{Tran-Thien_2023,LANTERN, approx} have explored relaxing the exact sampling constraints to further boost acceptance rates. Among these, LANTERN \cite{LANTERN} proposes enlarging the numerator in the acceptance criterion by incorporating neighboring tokens' probabilities, similar to our approach. However, this method has several issues, such as requiring training and lacking a detailed analysis of image AR acceptance rates. Critically, since it increases only the numerator, it shares the same drawbacks as \textit{naive lossy} methods—bias and the exploding problem—achieving only a 1.6× speed-up.

\label{sec:exp}

\section{Conclusion}
\label{sec:conc}
In this work, we propose \textbf{Grouped Speculative Decoding (GSD)}, a novel training-free SD method specifically designed for image AR. We first thoroughly analyze the primary challenge of SD in image AR and identify that the core issue arises from high-entropy predictions of the next token caused by the inherent variability of the images. Motivated by this, we introduce GSD, which effectively improves the acceptance rate of SD by smoothing the probability distributions through clustering semantically similar tokens. Furthermore, we observe that naive clustering relying on static embedding distances yields suboptimal outcomes, leading us to propose a dynamic clustering approach. Experimental results demonstrate that GSD achieves an average speed-up of \(3.8\times\) while maintaining image quality.

\paragraph{Acknowledgments}   
This work was supported by IITP and NRF grant funded by the Korea government(MSIT) (No. RS-2019-II191906, RS-2024-00415602) and Samsung Research Global AI Center.

{
    \small
    \bibliographystyle{ieeenat_fullname}
    \bibliography{main}
}

\clearpage
\maketitlesupplementary

\renewcommand{\thesection}{S\arabic{section}}
\renewcommand{\thefigure}{S\arabic{figure}}
\renewcommand{\thetable}{S\arabic{table}}

\renewcommand{\theequation}{S\arabic{equation}}

\setcounter{section}{0}
\setcounter{figure}{0}
\setcounter{table}{0}
\setcounter{equation}{0}
\section{Proof of proposition 1}

\begin{align*}
    \alpha_{p,q} &= \mathbb{E}_{q(x)}\left[\min\left(1, \frac{p(x)}{q(x)}\right)\right]\\
    &= \sum_x q(x)\min\left(1,\frac{p(x)}{q(x)}\right) \\
    &= \sum_x 
    \begin{cases}
        p(x), & \text{if } p(x) < q(x) \\[6pt]
        q(x), & \text{otherwise}
    \end{cases}\\
    &= \sum_x \min(p(x), q(x)) \\
    &= \sum_x \frac{p(x)+q(x)-|p(x) - q(x)|}{2}\\
    &= \frac{1}{2}\left(\sum_x p(x) + \sum_x q(x) - \sum_x|p(x) - q(x)|\right)\\
    &= \frac{1}{2}\left(1 + 1 - \sum_x|p(x) - q(x)|\right)\\
    &= 1-\frac{1}{2}\sum_x|p(x) - q(x)|\\[6pt]
    &= 1 - TV(p,q)
\end{align*}

\section{Full algorithm of GSD}

In this section, we provide detailed pseudocode for the implementation of GSD in Algorithm \ref{alg:gsd} and \ref{alg:gsdverify}. Since GSD is built upon SJD, it operates by simply replacing the VERIFY($\cdot$) part with our VERIFY\_GSD($\cdot$). For a more detailed implementation, please refer to the source code.

\begin{algorithm}
\caption{Speculative Jacobi Decoding\cite{SJD}}
\begin{algorithmic}[1]
\Require Speculative Length $L$, maximum seq length $N$, expert model $p_\theta $, initial context $X_{0:n_0 }$
\State $k \gets L$, $n \gets n_{0}$
\While{$n < N$}
\State $q_{L-k:L}, \hat{X}_{L-k:L} \sim \texttt{Rand-init}(\cdot) $
    \State \textbf{parallel for} $j = 0$ to $L$ \Comment{Parallel Verify}
        \State \quad $p_{j} \gets p_{\theta}(\cdot \mid [X_{0:n}, \hat{X}_{0 : j}])$
    \State \textbf{end for}
\State $(\hat{X}_{0:k},\, k) \gets \textcolor{blue}{\texttt{VERIFY}(\hat{X}_{0:L}, p_{0:L},\; q_{0:L})}$
    \State $X_{n:n+k-1} \gets \hat{X}_{0:k}$ \Comment{Accept.}
    \State $q_{0:L-k} \gets p_{k:L}$, $\hat{X}_{0:L-k} \gets \hat{X}_{k:L}$ \Comment{Draft update}\
    \State $n \gets n+k$

\EndWhile
\State \Return $X$
\end{algorithmic}
\label{alg:sjd}
\end{algorithm}

\begin{algorithm}
\caption{Grouped Speculative Decoding}
\label{alg:speculative}
\begin{algorithmic}[1]
\Require Speculative Length $L$, maximum seq length $N$, expert model $p_\theta $, initial context $X_{0:n_0 }$, Group size $G$, Embedding distance matrix $M_d$, thresholds $d,\delta$
\State $k \gets L$, $n \gets n_{0}$
\While{$n < N$}
\State $q_{L-k:L}, \hat{X}_{L-k:L} \sim \texttt{Rand-init}(\cdot) $
    \State \textbf{parallel for} $j = 0$ to $L$ \Comment{Parallel Verify}
        \State \quad $p_{j} \gets p_{\theta}(\cdot \mid [X_{0:n}, \hat{X}_{0 : j}])$
    \State \textbf{end for}
\State $(\hat{X}_{0:k},\, k) \gets \textcolor{red}{\texttt{VERIFY\_GSD}(\hat{X}_{0:L}, p_{0:L},\; q_{0:L},G)}$
    \State $X_{n:n+k-1} \gets \hat{X}_{0:k}$ \Comment{Accept.}
    \State $q_{0:L-k} \gets p_{k:L}$, $\hat{X}_{0:L-k} \gets \hat{X}_{k:L}$ \Comment{Draft update}\
    \State $n \gets n+k$

\EndWhile
\State \Return $X$
\end{algorithmic}
\label{alg:gsd}
\end{algorithm}

\section{Additional Results}
\begin{figure*}[ht!]
    \centering
    \includegraphics[width=\textwidth]{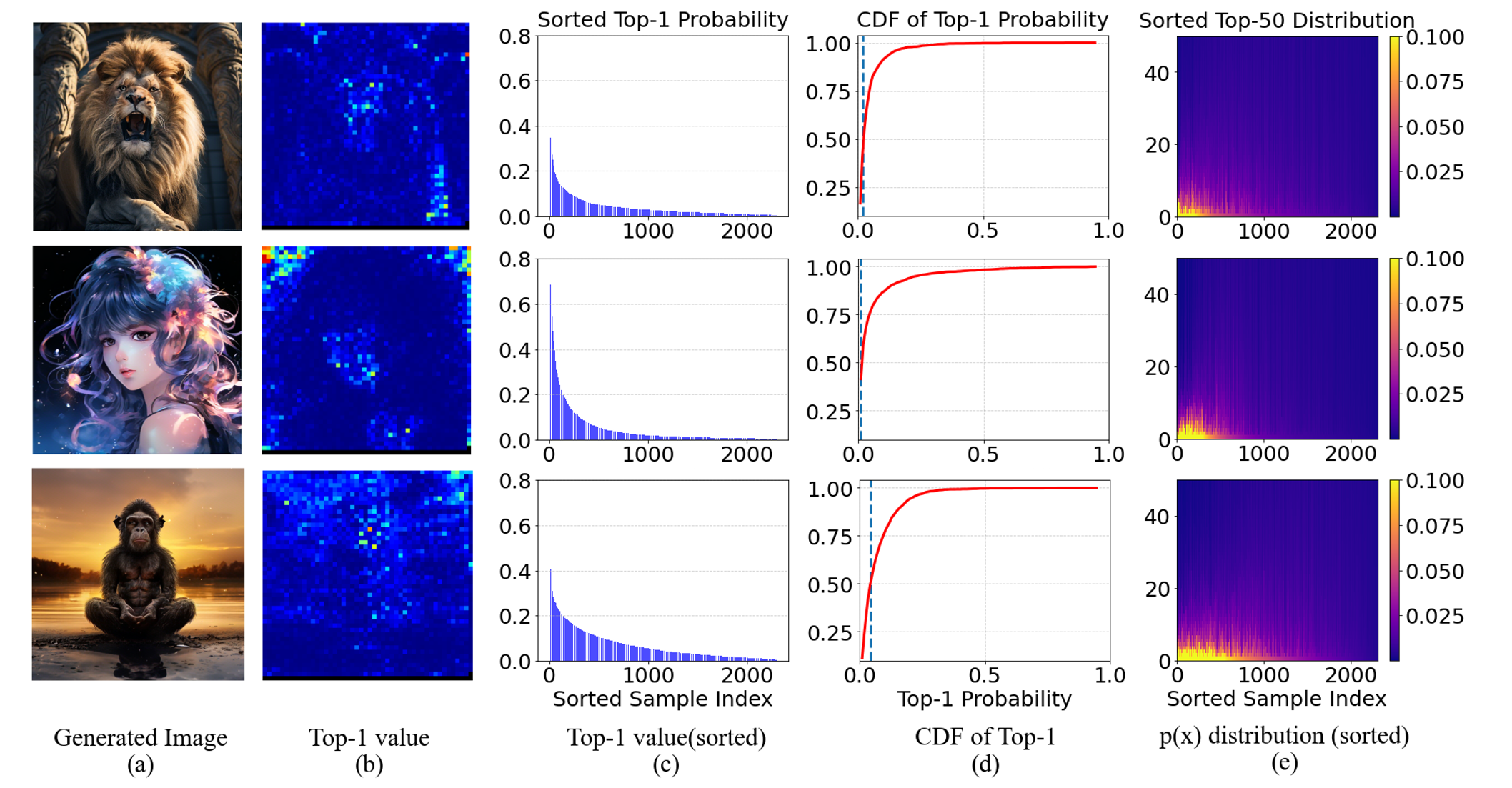}
    \caption{Additional p(x) visualization.}
    \label{fig:p_distribution_appendix}
\end{figure*}

In this section, we present additional experiments expanding upon the visualizations discussed in the main text. 
\paragraph{Top-1 probabilities } In Fig. \ref{fig:p_distribution_appendix}, we illustrate the visualization of Top-1 probabilities across a wider variety of images. As shown, regardless of the prompts, many images exhibit numerous tokens with low Top-1 probability distributions. 

\begin{figure*}[t!]
    \centering
    \includegraphics[width=0.99\textwidth]{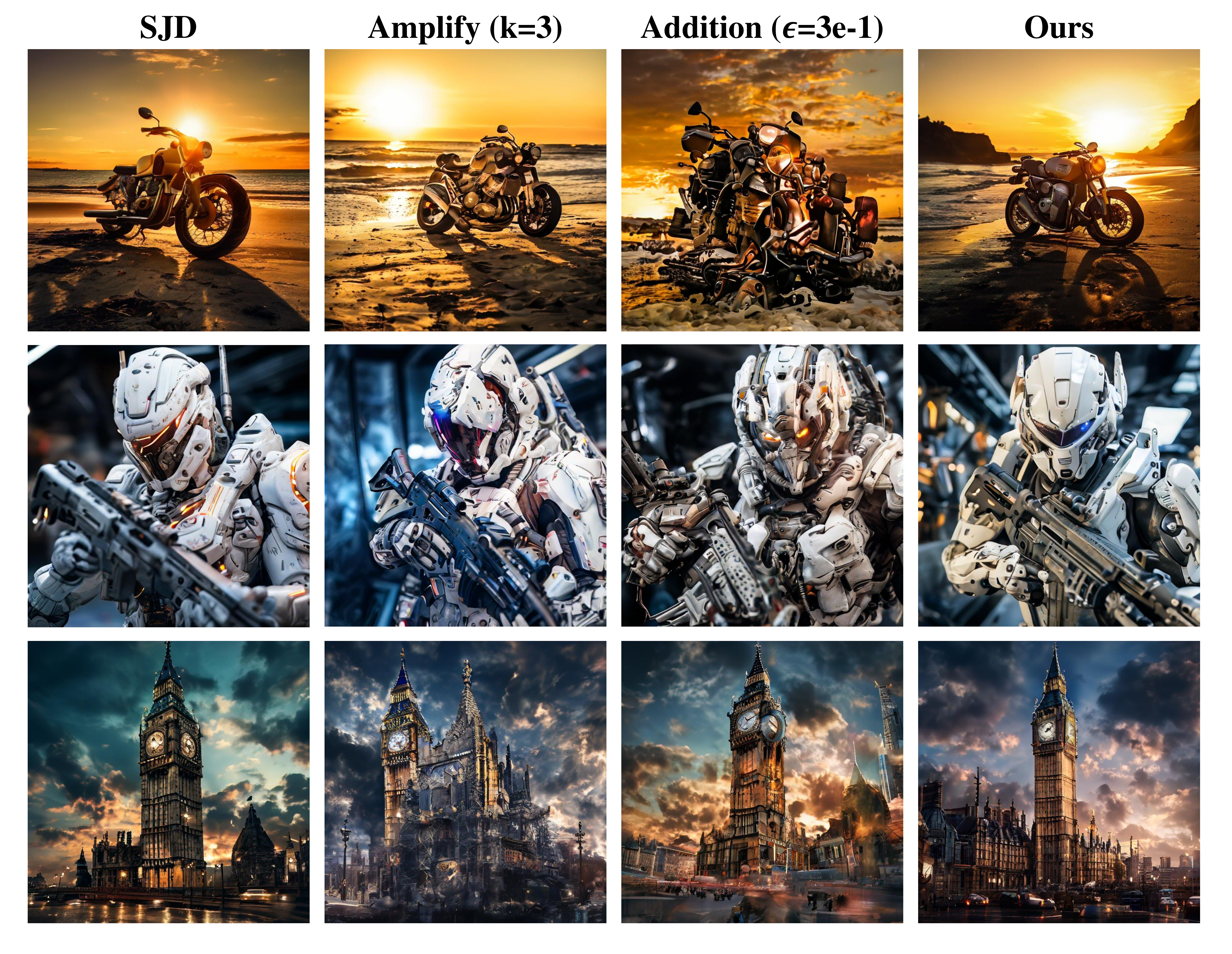} 
    \caption{Qualitative comparison between methods in Table 1 of the main paper}
    \label{fig:qual_comparison_appendix}
\end{figure*}

\paragraph{Visual quality comparison} In Fig. \ref{fig:qual_comparison_appendix}, we visually illustrate the differences in generation quality among various methods compared in Table 1. As shown in the figure, our GSD achieves approximately a 4× speed-up while maintaining generation quality comparable to lossless methods such as vanilla AR and SJD. In contrast, the naive lossy method also achieves acceleration but significantly degrades generation quality.

\begin{figure*}[t!]
    \centering
    \includegraphics[width=0.99\textwidth]{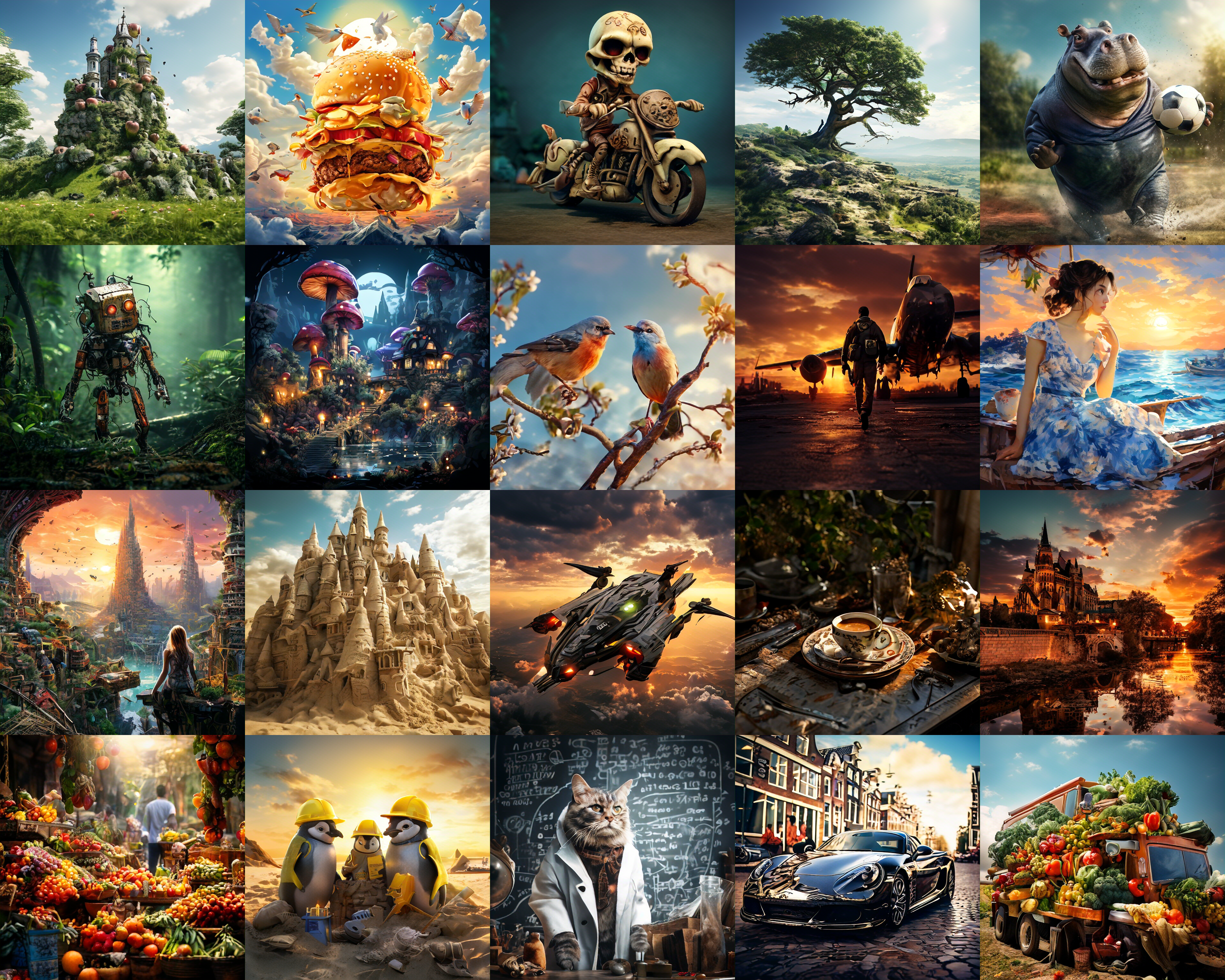} 
    \caption{Qualtitiave experiment on various prompt. Our GSD shows on average 3.6x NFE acceleration while maintaing image quality}
    \label{fig:qual_appendix}
\end{figure*}

\paragraph{GSD generation performance} Fig. \ref{fig:qual_appendix} presents further qualitative results of our method when accelerated by an average factor of 3.6. As demonstrated in the figure, our GSD significantly accelerates AR image decoding while maintaining generation quality across diverse prompts. 

\section{Prompts on Qualititave Experiment}

In Figure. 9 on main paper, the prompts for each images are as follows :

\begin{itemize}
    \item \textit{Rusty robot on a skateboard in the hallway of domitory, photography, 4k, realistic, detailed, bright}
    \item \textit{Origami astronaut, waliking in the cloud, bright background, realistic, 4k, photography, bright color}
    \item \textit{photography, realistic, White cute fluffy dog, skyblue background, very intricate, very detailed, realistic., bright}
    \item \textit{color photo, photography, Face of a young man, very detailed, realistic. sharp, film grain, high contrast}
    \item \textit{animation art work, cute, cat character, bright color pallette}

\end{itemize}

\end{document}